# MEBN-RM: A Mapping between Multi-Entity Bayesian Network and Relational Model

*Cheol Young Park*[a]* and *Kathryn Blackmond Laskey*[a]

[a]*The Sensor Fusion Lab & Center of Excellence in C4I, George Mason University, USA*

A B S T R A C T

*Multi-Entity Bayesian Network* (MEBN) is a knowledge representation formalism combining *Bayesian Networks* (BN) with *First-Order Logic* (FOL). MEBN has sufficient expressive power for general-purpose knowledge representation and reasoning. Developing a MEBN model to support a given application is a challenge, requiring definition of entities, relationships, random variables, conditional dependence relationships, and probability distributions. When available, data can be invaluable both to improve performance and to streamline development. By far the most common format for available data is the *relational database* (RDB). Relational databases describe and organize data according to the *Relational Model* (RM). Developing a MEBN model from data stored in an RDB therefore requires mapping between the two formalisms. This paper presents MEBN-RM, a set of mapping rules between key elements of MEBN and RM. We identify links between the two languages (RM and MEBN) and define four levels of mapping from elements of RM to elements of MEBN. These definitions are implemented in the MEBN-RM algorithm, which converts a relational schema in RM to a partial MEBN model. Through this research, the software has been released as a MEBN-RM open-source software tool. The method is illustrated through two example use cases using MEBN-RM to develop MEBN models: a Critical Infrastructure Defense System and a Smart Manufacturing System.

## 1. Introduction

Statistical relational learning(SRL) deals with representation and reasoning methods for uncertain and complex situations by combining probabilistic models (e.g., Bayesian Networks and Markov Networks) and relational structures (e.g., First-Order Logic and Relational Model) [Getoor & Taskar, 2007]. The expressive power of SRL enables us to represent real-world situations characterized by uncertainty and complexity. For this reason, it has been used in several domains (e.g., information fusion [Müller et al., 2017], video analysis [Morariu & Davis, 2011][Wu & Aghajan, 2011], and bioinformatics [Lippi & Frasconi, 2009]). Several formalisms embodying probabilistic models with relational structures have been proposed in the past decades, such as Probabilistic-Logic Programming [Poole, 1993], Programming in Statistical Modeling [Sato & Kameya, 1997], Probabilistic Relational Models [Koller, 1999], Relational Bayesian Networks [Jaeger, 1997], Relational Markov Networks [Taskar et al., 2002], Bayesian LOGic [Milch et al., 2005], Markov Logic Networks [Richardson & Domingos, 2006], Conditional Random Fields for Logical Sequences [Gutmann & Kersting, 2006], Bayesian Logic Programming [Kersting & De Raedt, 2007], FACTORIE: Probabilistic Programming [McCallum et al., 2009], and Probabilistic Conditional Logic [Beierle et al., 2017].

Multi-Entity Bayesian Network (MEBN) belongs to the formalisms in SRL. MEBN is a knowledge representation language based on Bayesian Networks (BN) [Pearl, 1988] and First-Order Logic (FOL). MEBN is sufficiently expressive for general-purpose knowledge representation and reasoning in an uncertain and complex world. Because MEBN is flexible enough to represent a variety of complex and uncertain situations, it has been applied to systems for *Predictive Situation Awareness* (PSAW), the problem of understanding and predicting aspects of a temporally evolving situation [Laskey et al., 2000][Wright et al., 2002][Suzic, 2005][Costa et al., 2009][Carvalho et al., 2010][Park et al., 2014][Golestan, 2016][Li et al., 2017][Park et al., 2017]. In a recent review of knowledge representation formalisms, Golestan et al. [2016] recommended MEBN as having the most comprehensive coverage of features needed to represent complex problems among several artificial intelligence (AI) models including statistical relational models (e.g., Probabilistic Relational Models [Getoor & Taskar, 2007] and Markov Logic Networks [Richardson & Domingos, 2006]).

* *Corresponding author.*
E-mail address: cparkf@gmu.edu

Construction of relational structures (e.g., First-Order Logic and Relational Model) for MEBN is an active research topic. Probabilistic Web Ontology Language (PR-OWL) [Costa, 2005] extends the Web Ontology Language (OWL) to represent uncertainty. A domain ontology, typically represented in a relational language, provides common semantics for expressing information about entities and relationships in a domain. PR-OWL is an upper ontology, written in OWL, that uses MEBN to express uncertainty about entities and relationships in an OWL ontology. PR-OWL 2 extends PR-OWL to provide better integration with OWL [Carvalho et al., 2017]. Carvalho et al. [2016] provides a methodology for developing MEBN theories expressed as PR-OWL ontologies.

This paper focuses on *Relational Model*, which is another kind of relational structure and from which MEBN theories can be partially constructed. The Relational Model (RM) [Codd, 1969; Codd, 1970] is the most popular database model. While non-relational databases, called *NoSQL*, are receiving increasing attention [Han et al., 2011], our focus in this work is on RM because so much of the available data is stored in relational databases. A *Relational Database* (RDB) uses RM to describe and organize data. An RDB stores data in the form of multiple *relations*. A relation is composed of a *relation schema* and a *relation instance*. The relation schema represents a class of entities and its attributes. A *relational database schema* or *relational schema* is a collection of relation schemas. A MEBN model, called an *MTheory*, consists of a set of *MFrags*. An MFrag is composed of *Context nodes*, *Input nodes*, *Resident nodes*, a fragment *Graph*, and a set of *Local Distributions*.

In order to construct an MTheory from an RDB, we need a way to map from relations to MFrags. In this paper, we introduce a mapping between a relational schema and a partial MTheory. This mapping is called *MEBN-RM Mapping* (or MEBN-RM). MEBN-RM contains four levels of mapping from elements of a relational database to elements of an MTheory. The first level maps a relation schema to an entity in an MTheory. The second level maps attributes of a relation schema to resident nodes of an MFrag. The third maps a relation schema to an MFrag is defined. The fourth level maps a relational database to an MTheory. Further, MEBN-RM forms the basis for a *MEBN-RM mapping algorithm* takes a relational database as input and produces a partial MTheory as output.

The remainder of the paper is organized as follows. Section 2 provides background knowledge on MEBN and RM. Section 3 defines MEBN-RM and presents the mapping algorithm. Section 4 introduces a MEBN-RM open-source software tool and an experiment for MEBN-RM algorithm performance in terms of the mapping time and accuracy. Section 5 presents two use cases in which the tool is applied to construct a partial MTheory. The final section presents conclusions and future research directions.

## 2. MEBN-RM

Both MEBN and RM have their theoretical basis in first-order logic, and both represent entities in a domain and relationships among them.

## 3. Background

In this section, we describe MEBN, a graphical representation for MEBN, and a script form of MEBN. Then, RM is presented briefly along with examples.

### *3.1. Multi-Entity Bayesian Network*

MEBN [Laskey, 2008] allows compact representation of repeated structure in a joint distribution on a set of random variables. In MEBN, random variables are defined as templates that can be repeatedly instantiated to construct probabilistic models with repeated structure. MEBN represents domain knowledge using an MTheory, which consists of a collection of MFrags (see Fig. 1). An MFrag is a fragment of a graphical model that is a template for probabilistic relationships among instances of its random variables. Random variables (RVs) may contain ordinary variables, which can be instantiated for different domain entities. We can think of an MFrag as a class which can generate instances of BN fragments. These can then be assembled into a Bayesian network, called a *situation-specific Bayesian Network* (SSBN), using an SSBN algorithm [Laskey, 2008]. A given MTheory can be used to construct many different SSBNs for different situations.

To understand how this works, consider Fig. 1, which shows an MTheory called the *Danger Assessment* MTheory. This MTheory contains seven MFrags: *Speed*, *ImageTypeReport*, *VehicleObject*, *Danger*, *Weather*, *Region*, and *Reference*. An MFrag may contain three types of random variables: *context* RVs, denoted by green pentagons, *resident* RVs, denoted by yellow ovals, and *input* RVs, denoted by gray trapezoids. Each MFrag defines local probability distributions for its input RVs. These distributions may depend on the input RVs, whose distributions are defined in other MFrags. Context RVs express conditions that must be satisfied for the distributions defined in the MFrag to apply.

For example, consider the *VehicleObject* MFrag in the MTheory of Fig. 1. This MFrag expresses knowledge of how the vehicle class is related to the terrain type. The context RVs, *IsA*(*obj*, VEHICLE), *IsA*(*rgn*, REGION), and *rgn* = *Location*(*obj*), indicate that the ordinary variable *obj* must refer to a vehicle, the ordinary variable *rgn* must refer to a region, and the object denoted by *obj* must be located in the region denoted by *rgn*. The resident RV *VehicleClass*(*obj*) refers to the type of *obj*. This type is uncertain, with its distribution defined in the MFrag. The distribution depends on the type of terrain in the region, with wheeled vehicles more likely on roads and tracked vehicles more likely on rough terrain. Terrain type is represented by the input RV *TerrainType*(*rgn*), whose distribution is defined in the *Region* MFrag.

The distribution for a resident random variable, defined in its home MFrag, is called a *class local distribution*. For example, the class local distribution (CLD) of *VehicleClass*(*obj*), which depends on the type of terrain where the vehicle is located, can be expressed as CLD 1.

**CLD1**: The class local distribution for the *VehicleClass*(*obj*) RV

```
if some rgn have (TerrainType = Road) [
    Tracked = .2,
    Wheeled = .8
] else if some rgn have (TerrainType = OffRoad) [
    Tracked = .8,
    Wheeled = .2
] else [
    Tracked = .5,
    Wheeled = .5
```

This class local distribution specifies that if the terrain type is *Road*, then there is an 80% chance the vehicle is wheeled; and if the terrain type is *OffRoad*, then there is an 80% chance the vehicle is tracked. The final clause specifies a distribution if none of these conditions is met, which is equal probabilities for tracked and wheeled vehicles.

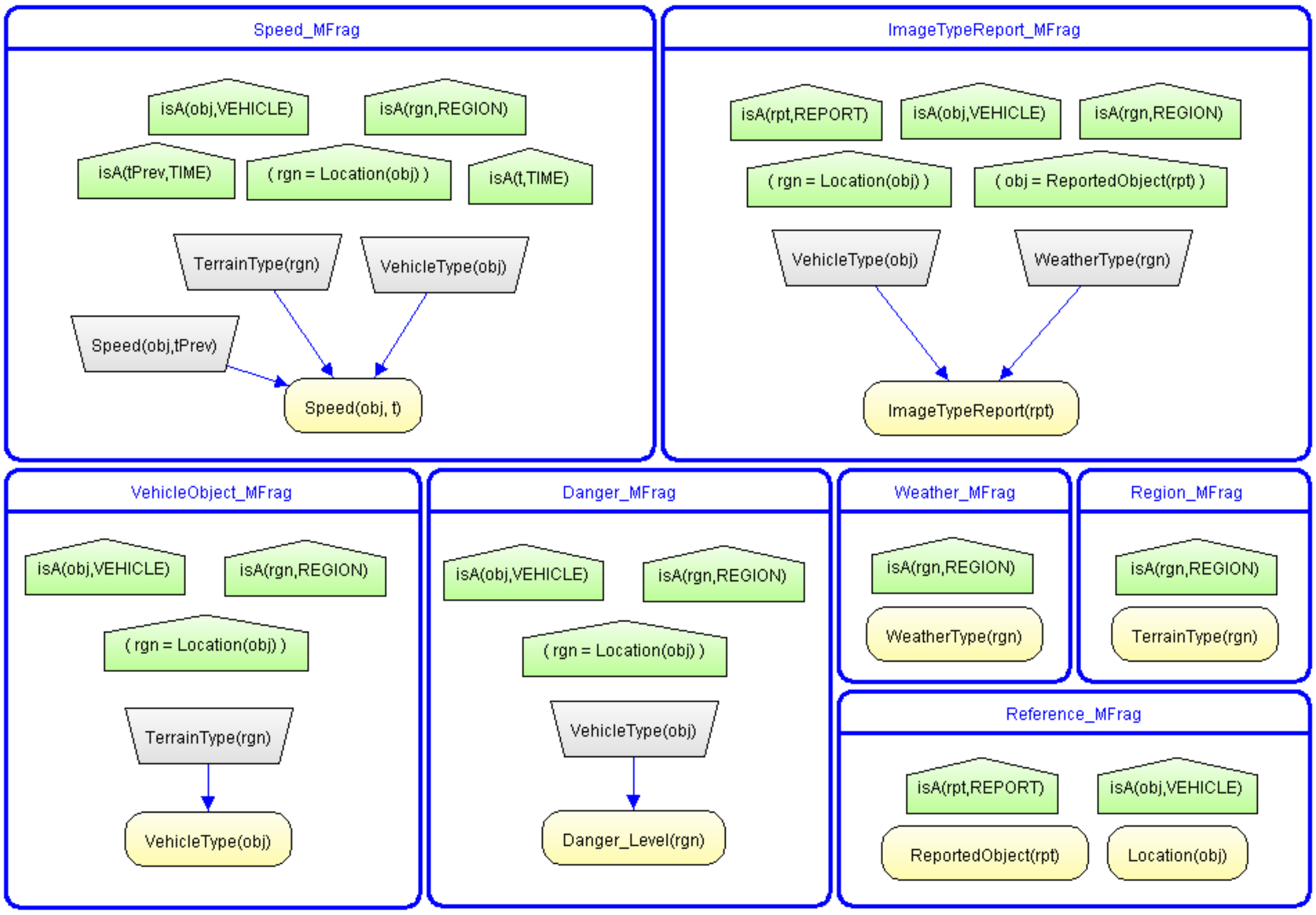

**Figure 1.Danger Assessment MTheory**

Formally, an MFrag is defined as follows [Laskey, 2008].

**Definition 2.1 (MFrag)** An MFrag *F*, or MEBN fragment, consists of: (*i*) a set ***C*** of context nodes, which represent conditions under which the distribution defined in the MFrag is valid; (*ii*) a set ***I*** of input nodes, which have their distributions defined elsewhere and condition the distributions defined in the MFrag; (*iii*) a set ***R*** of resident nodes, whose distributions are defined in the MFrag[*]; (*iv*) an acyclic directed graph ***G***, whose nodes are associated with resident and input nodes; and (*iv*) a set $\boldsymbol{L}^C$ of class local distributions, in which an element of $\boldsymbol{L}^C$ is associated with each resident node.

[*] Bold italic letters are used to denote sets.

The nodes in an MFrag are different from the nodes in a common Bayesian network. A node in a common BN represents a single random variable, whereas a node in an MFrag represents a collection of RVs: those formed by replacing the ordinary variables with identifiers of entity instances that meet the context conditions. To emphasize the distinction, we call the resident nodes in the MFrag MEBN nodes, or MNodes.

MNodes correspond to predicates (for true/false RVs) or terms (for other RVs) of first-order logic. An MNode is written as a predicate or term followed by a parenthesized list of ordinary variables as arguments.

**Definition 2.2 (MNode)** An *MNode*, or MEBN Node, is a random variable *N*(*ff*) corresponding to an *n*-ary function or predicate of first-order logic, a list of *n* arguments consisting of ordinary variables, a set of mutually exclusive and collectively exhaustive possible values, and an associated class local distribution. The special values *true* and *false* are the possible values for predicates, but may not be possible values for functions. The RVs associated with the MNode are constructed by substituting domain entities for the *n* arguments of the function or predicate. The class local distribution specifies how to define local distributions for these RVs.

For example, the node *VehicleClass*(*obj*) in Fig. 1 is an MNode corresponding to the FOL function *VehicleClass*(*obj*). It has two possible values (i.e., *Wheeled* and *Tracked*). This MNode is associated with the class local distribution, $\boldsymbol{L}^C$ in CLD 1. The MNode is used as a template for the distributions of instance RVs created when an SSBN is constructed from the MFrag associated with the MNode. These instances are formed by substituting identifiers of vehicle objects for the ordinary variable *obj*.

**Definition 2.3 (MTheory)** An *MTheory M*, or MEBN Theory, is a collection of MFrags that satisfies conditions given in [Laskey, 2008] ensuring the existence of a unique joint distribution over its random variables.

An MTheory is a collection of MFrags that defines a consistent joint distribution over random variables describing a domain. The MFrags forming an MTheory should be mutually consistent. To ensure consistency, conditions must be satisfied such as no-cycle, bounded causal depth, unique home MFrags, and recursive specification condition [Laskey, 2008]. No-cycle means that the generated SSBN will contain no directed cycles. Bounded causal depth means that depth from a root node to a leaf node of an instance SSBN should be finite. Unique home MFrags means that each random variable has its distribution defined in a single MFrag, called its home MFrag. Recursive specification means that MEBN provides a means for defining the distribution for an RV depending on an ordered ordinary variable from previous instances of the RV.

The *IsA* random variable is a special RV representing the type of an entity. IsA is commonly used as a context node to specify the type of entity that can be substituted for an ordinary variable in an MNode.

**Definition 2.4 (IsA Random Variable)** An *IsA random variable*, IsA(*ov*, *tp*), is an RV corresponding to a 2-argument FOL predicate. The IsA RV has value *true* when its second argument *tp* is filled by the type of its first argument *ov* and *false* otherwise.

For example, in the *Danger* MFrag in Fig. 1, *IsA*(*obj*, VEHICLE) is an *IsA* RV. Its first argument *obj* is filled by an entity instance and its second argument is the type symbol Vehicle. It has value *true* when its first argument is filled by an object of type Vehicle.

MEBN is a highly expressive language, combining the expressiveness of first-order logic with a mathematically sound uncertainty representation. In a review of a number of knowledge representation formalisms, Golestan et al [2016] recommended MEBN as having the most comprehensive coverage. The reviewed formalisms included MEBN, Hidden Markov Models [Baum & Petrie, 1966], Artificial Neural Networks [Hopfield, 1988], Bayesian Networks [Pearl, 1988], Support Vector Machine [Cortes & Vapnik, 1995], Fuzzy Bayesian Networks [Pan & Liu, 2000], Dynamic Bayesian Networks [Murphy & Russell, 2002], and Markov Logic Networks [Richardson & Domingos, 2006]. MEBN has been applied to a wide variety of domains [Patnaikuni et al., 2017].

### *3.2. A Script for MEBN*

Fig. 1 shows a graphical representation for an MTheory. In this subsection, we introduce a script representing an MTheory. This script is useful to manage contents of an MTheory. The *Danger Assessment* MTheory in Fig. 1 can be represented by the following script (MTheory 1).

**MTheory 1**: Part of Script MTheory for Danger Assessment

```
1   [F: ImageTypeReport
2       [C: IsA(obj, VEHICLE)][C: IsA(rgn, REGION)][C: IsA(rpt, REPORT)]
3       [C: rgn = Location (obj)][C: obj = ReportedObject (rpt)]
4       [R: ImageTypeReport (rpt)
5               [IP: WeatherType (rgn)]
6               [IP: VehicleClass (obj)]
7       ]
8   ]
9   [F: Weather
10      [C: IsA(rgn, REGION)]
11      [R: WeatherType (rgn)]
12  ]
13  …
```

The script contains several predefined single letters (F, C, R, IP, RP, and L). The single letters, F, C, and R denote an MFrag, a context node, and a resident node, respectively. For a resident node (e.g., Y) in an MFrag, a resident parent (RP) node (e.g., X), which is defined in the MFrag, is denoted as RP (e.g., [R: Y [RP: X]]). For an input node, we use a single letter IP. Each node can contain a CLD denoted as L. For example, suppose that there is a CLD type called *WeatherCLD*. If the resident node *WeatherType* in Line 11 uses the CLD type *WeatherCLD*, the resident node *WeatherType* can be represented as [R: WeatherType (*rgn*) [L: *WeatherCLD*]].

### *3.3. Relational Model*

In 1969, Edgar F. Codd proposed the *Relational Model* (RM) as a database model based on first-order predicate logic [Codd, 1969][Codd, 1970]. The RM is the most popular database model. A *relational database* (RDB) is a database that uses the RM to describe and organize data. In the RM, data are organized as a collection of relations. A *relation* is an abstract definition of a class of entities or a relationship that can hold between classes of entities. An *instance of a relation* is depicted as a table in which each column is an attribute of the relation and each row, also called a *tuple*, contains the value of each attribute for an individual entity of the class represented by the relation. An entry in the table, called a *cell*, is the value of the attribute associated with the column for the entity associated with the row. A *key* for a relation is one or more attributes that uniquely identify a particular domain entity or row. A *primary key* uniquely identifies the individual entities in the relation. A *foreign key* points to the primary key in another relation. The *cardinality* of a relation is the number of rows in the table, i.e., the number of unique entities of the type represented by the relation. The *degree* of the relation is the number of columns in the table, i.e., the number of attributes of entities of the type represented by the relation.

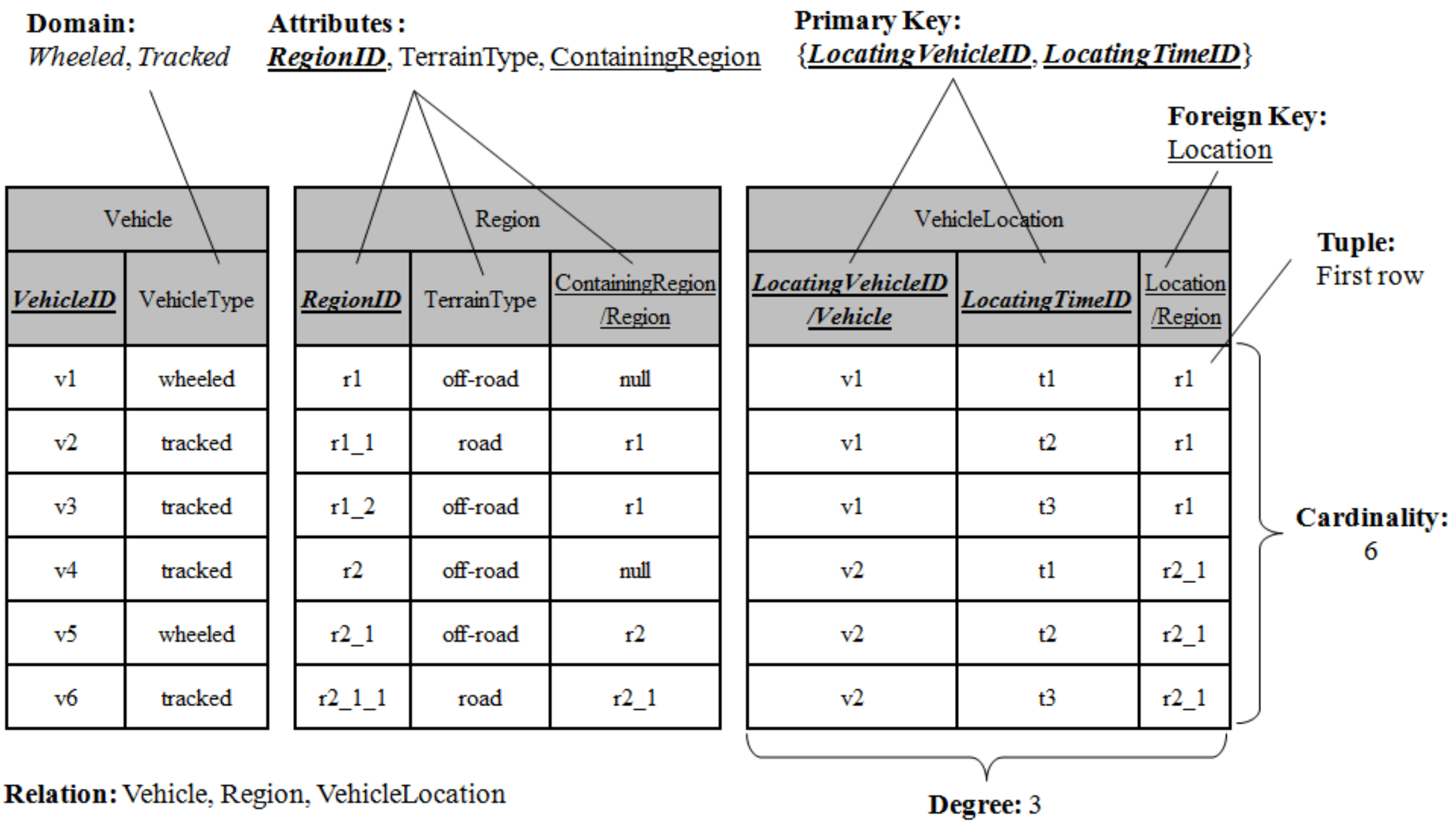

| Vehicle | |
|---|---|
| ***VehicleID*** | VehicleType |
| v1 | wheeled |
| v2 | tracked |
| v3 | tracked |
| v4 | tracked |
| v5 | wheeled |
| v6 | tracked |

| Region | | |
|---|---|---|
| ***RegionID*** | TerrainType | ContainingRegion /Region |
| r1 | off-road | null |
| r1_1 | road | r1 |
| r1_2 | off-road | r1 |
| r2 | off-road | null |
| r2_1 | off-road | r2 |
| r2_1_1 | road | r2_1 |

| VehicleLocation | | |
|---|---|---|
| ***LocatingVehicleID /Vehicle*** | ***LocatingTimeID*** | Location /Region |
| v1 | t1 | r1 |
| v1 | t2 | r1 |
| v1 | t3 | r1 |
| v2 | t1 | r2_1 |
| v2 | t2 | r2_1 |
| v2 | t3 | r2_1 |

**Figure 2. Example of a Part of the Vehicle Identification RDB**

Fig. 2 shows an illustrative example of an RDB. In the example RDB, there are three relations: *Vehicle*, *Region*, and *VehicleLocation*. We could imagine different situations, each with different vehicles, regions, etc. Each particular situation, like the one depicted in Fig. 2, corresponds to an instance of this relational model. The instance is represented as a table for each of the relations as shown Fig. 2, where the columns represent attributes of the relation and the rows represent the attribute values for specific entities. For example, the *Vehicle* relation has two attributes: *VehicleID*, which uniquely identifies each individual vehicle, and *VehicleClass*, which indicates whether the vehicle is tracked or wheeled. The *VehicleLocation* relation has three attributes: *LocatingVehicleID*, *LocatingTimeID*, and *Location*. The *LocatingVehicleID* attribute in the *VehicleLocation* relation is a foreign key pointing to the primary key of the *Vehicle* relation. A row of the *VehicleLocation* relation represents a vehicle being located in a region at a point in time. Attributes that are part of the primary key of the relation (e.g., ***LocatingVehicleID*** and ***LocatingTimeID*** in the *Location* relation) are denoted by bold, italicized, and underlined letters, while foreign keys which are not part of the primary key of a relation in which the foreign keys are used (e.g., Location in the *VehicleLocation* relation) are denoted by underlined letters. A relation without instances or data – that is, an empty table – is called the *relation schema*.

**Definition 2.5 (Key)** A *key* of a relation schema is a set of one or more attributes that uniquely identify a row of the relation.

**Definition 2.6 (Foreign Key)** A *foreign key*, FK, of a relation schema, RS[$A_1$, $A_2$, ..., $A_n$], is a subset of the attributes {$A_1$, $A_2$,..., $A_n$} that uniquely identifies a row of another relation.

The *VehicleLocation* relation of Fig. 2 has two foreign keys (i.e., ***LocatingVehicleID/Vehicle*** and Location/Region). Here, we use the "/" symbol followed by the relation name to indicate the relation to which the foreign key points, i.e., ***LocatingVehicleID*** foreign key refers to the *Vehicle* relation. A relation schema containing a foreign key is called a *target relation schema* for the foreign key, while a relation schema which is referenced by the foreign key is called a *home relation schema*. For example, the *VehicleLocation* relation schema is the target relation schema for the ***LocatingVehicleID*** attribute, while the *Vehicle* relation schema is the home relation schema of the attribute. If the *target and home relation schema* are same, the foreign key and primary key are the same. In this case, the foreign key is called a *recursive foreign key*.

**Definition 2.7 (Primary Key)** A *primary key*, PK, of a relation schema RS[$A_1$, $A_2$, ..., $A_n$] is a selected subset of the attributes {$A_1$, $A_2$,..., $A_n$} that uniquely identifies each tuple in the RS.

The *VehicleLocation* relation of Fig. 2 has a primary key composed of two attributes (i.e., ***LocatingVehicleID/Vehicle*** and ***LocatingTimeID***). Each tuple of the *VehicleLocation* relation is uniquely identified by these two arguments. Note that the attribute Location/Region in the relation is not used as the primary key in the target relation, but it uniquely identifies each tuple in its home relation (i.e., the *Region* relation); therefore, it is a foreign key. This kind of key is called a *Non-Primary Foreign Key*.

**Definition 2.8 (Non-Primary Foreign Key)** A *Non-Primary Foreign Key*, NK, is a Foreign Key that is not used for a primary key in a target relation.

The *ContainingRegion* attribute of the *Region* relation of Fig. 2 is another example of a non-primary foreign key since the home relation schema of the *ContainingRegion* attribute is the *Region* relation and it is not used for the primary key of the target relation.

**Definition 2.9 (Non-Foreign-Key Attribute)** A *Non-Foreign-Key Attribute*, A, is an attribute which is not a foreign key.

For example, in Fig. 2, the *VehicleClass* and *TerrainType* attribute are non-foreign-key attributes since they are not foreign keys.

**Definition 2.10 (Original Primary Key)** An *Original Primary Key*, OK, is a primary key that is not a foreign key in any target relation.

A primary key in one relation can be imported from the primary key in another relation (i.e., foreign & primary key), while an original primary key is not originated from other primary key in another relation. Examples for the original primary key can include ***VehicleID*** in the *Vehicle*relation, ***RegionID*** in the *Region* relation, and ***LocatingTimeID*** in the *Location* relation.

**Definition 2.11 (Relation Schema)** A *relation schema*, RS[$A_1$:$D_1$, $A_2$:$D_2$, ..., $A_n$:$D_n$], is a set of pairs $A_i$:$D_i$, where $A_1 \neq \ldots \neq A_n$ are *attribute names*, and $D_i$ is a set called the *domain* for attribute i.

For example, the relation schema of *Vehicle* in Fig. 2 is [***VehicleID***:{v1, v2, …}, VehicleClass:{wheeled, tracked}], where ***VehicleID*** and VehicleClass are attributes with domains {v1, v2, …} and {wheeled, tracked}, respectively. Note that we denote the domain of the attribute by inserting the colon, ":", between the name of the attribute and the name of the domain. This ancillary information can be omitted for brevity (e.g., [***VehicleID***, VehicleClass]). As another example, the relation schema of the *VehicleLocation* relation is VehicleLocation[***LocatingVehicleID***:{v1, v2, …}, ***LocatingTimeID***:{t1, t2, …}, Location:{r1, r1_1, r1_2, …}], where ***LocatingVehicleID***, ***LocatingTimeID***, and Location are attributes, with domains {v1, v2, …}, {t1, t2, …}, and {r1, r1_1, r1_2, …}, respectively.

**Definition 2.12 (Entity Relation Schema)** An entity relation schema, ERS, is a relation schema containing an original primary key that is not a foreign key and that consists of exactly one attribute.

In Fig. 2, the *Vehicle* and *Region* relation schema are entity relation schemas. An entity relation schema represents a type of entity. The original primary key is a field that holds an identifier that uniquely identifies an instance of the entity type.

**Definition 2.13 (Relationship Relation Schema)** A relationship relation schema, RRS, is a relation schema containing a primary key consisting of attributes which are foreign keys pointing to entity relation schemas.

Therefore, a relationship relation represents a relationship for one entity type (i.e., a unary relation) or a relationship between entities of more than two entity types. In Fig. 2, the *VehicleLocation* relation schema is a relationship relation schema, if we assume that there is a *Time* entity relation schema and the *LocatingTimeID* attribute points to the *Time* entity relation schema.

**Definition 2.14 (Relation Instance)** A *relation instance*, RI of a relation schema, RS[$A_1$, $A_2$, ..., $A_n$], is specified by a table with *n* columns and *m* rows, {{$d_{11}$, $d_{21}$, ..., $d_{n1}$},... {$d_{1m}$, $d_{2m}$, ..., $d_{nm}$}}, where $A_i$ is an attribute of RS, $d_{ij} \in$ Dom($A_i$)[†]. The relation instance represents a set of *m* specific entities of the class represented by the relation.

For example, in the *Vehicle* relation, there are six rows (i.e., {{v1, wheeled}, {v2, tracked}, {v3, tracked}, {v4, tracked}, {v5, wheeled}, {v6, tracked}}). The instance of the *Vehicle* relation refers to these six rows, or tuples.

**Definition 2.15 (Relational Database Schema)** A *relational database schema,* RDBS[$RS_1$, $RS_2$, ..., $RS_n$], is a set $RS_i$ of relation schemas.

For example, the table headers of Fig. 2 describe the relational database schema, RDBS[*Vehicle*, *Region*, *VehicleLocation*].

**Definition 2.16 (Relational Database)** A *relational database*, RDB[$RI_1$:$RS_1$, $RI_2$:$RS_2$, …, $RI_n$:$RS_n$], is a set of pairs $RI_i$:$RS_i$, where $RS_i$ denotes a relation schema and $RI_i$ denotes an instance of $RS_i$.

For example, the tables of Fig. 2 describe the relational database RDB[{{v1, wheeled}…{v6, tracked}}:*Vehicle*, {{r1, off-road, null}…{r2_1_1, road, r2_1}}:*Region*, {{v1, t1, r1}…{v2, t3, r2_1 }}:*VehicleLocation*].

In the relational model, *normalization* is an operation performed on an RDB to make it more manageable by minimizing redundancy of elements and reducing dependency between attributes [Codd, 1970]. Several normal forms have been suggested such as First ~ Fifth normal form and Boyce–Codd Normal Form (BCNF) [Codd, 1970][Codd, 1972][Codd, 1974][Fagin, 1977][Fagin, 1979][Maier, 1983].

## 4. MEBN-RM

Both MEBN and RM have their theoretical basis in first-order logic, and both represent entities in a domain and relationships among them.

## 5. MEBN-RM

Both MEBN and RM have their theoretical basis in first-order logic, and both represent entities in a domain and relationships among them. We would like to be able to use data stored in a RDB to learn the joint distribution represented by an MTheory. To do this, we need a mapping from elements of RM to elements of MEBN. MEBN-RM provides such a mapping. MEBN-RM contains four levels of mapping from elements of a relational database to elements of an MTheory. In the first level, an entity mapping between an entity relation schema and an entity is defined. In the second level, a resident node mapping is defined. In the third level, a relation and MFrag mapping is defined. In the fourth level, a relational database schema and MTheory mapping is defined. Before discussing these mappings, some ingredients and assumptions are discussed in this section.

The following RDBS 1 from the *Vehicle Identification* RDB in Section 2 is used for an illustrative example through Section 3. In the example RDBS, there are four relations: *Vehicle*, *Region*, *VehicleLocation*, and *Follow*. The relation *Vehicle* has two attributes: ***VehicleID*** and VehicleClass. The relation *Region* has three attributes: ***RegionID***, TerrainType, and ContainingRegion. The relation *VehicleLocation* has three attributes:

[†] *Dom(X) is the domain of the attribute X. This is the set of values that the attribute can take on.*

***LocatingVehicleID***, ***LocatingTimeID***, and Location. The relation *Follow* has two attributes: ***FollowingVehicleID/Vehicle*** and ***LeadingVehicleID/Vehicle***.

RDBS 1 [Vehicle Identification]

```
VehicleIdentification[
 Vehicle[VehicleID, VehicleClass],
 Region[RegionID, TerrainType, ContainingRegion/Region],
 VehicleLocation[LocatingVehicleID/Vehicle, LocatingTimeID, Location/Region],
 Follow [FollowingVehicleID/Vehicle, LeadingVehicleID/Vehicle]
 ]
```

As we saw in Section 2, an attribute in a relation can be a *Primary key* (PK), *Non-Foreign-Key Attribute* (NF), or *Non-Primary Foreign Key* (NK). For example, in the relation *Region*, the attribute ***RegionID*** is PK, the attribute TerrainType is NF, and the attribute ContainingRegion is NK. Because each of these types of attribute plays a different role in MEBN-RM, we distinguish them from each other.

In Section 2, Definition 2.11 defined a relation schema as a set of pairs consisting of an attribute name and a domain (i.e., RS[$A_1$:$D_1$, $A_2$:$D_2$, ..., $A_n$:$D_n$], where $A_i$ is the *i*-th attribute name and $D_i$ is the domain for the attribute *i*). The attributes[$A_1$, $A_2$, ..., $A_n$] can be grouped into three disjoint and exhaustive subsets: PK, NF, and NK, where PK is the set of attributes in a primary key, NF is the set of non-foreign-key attributes, and NK is the set of attributes in a non-primary foreign key.

A variety of relations can be formed in accordance with the following restrictions on the attributes in these subsets. A PK in a relation cannot be empty; however an NF, NK, or both in a relation may be empty. Therefore, a relation can be one of four types: (1) RS[PK] denotes a relation schema containing only a primary key, (2) RS[PK, NF] denotes a relation schema containing only a primary key and non-foreign-key attributes, (3) RS[PK, NK] denotes a relation schema containing only a primary key and non-primary foreign key attributes, and (4) RS[PK, NF, NK] denotes a relation schema containing a primary key, non-foreign-key attributes, and non-primary foreign key attributes. We define the mapping between RM and MEBN for each of these four types of relation.

We start by assuming that all relations in the RDB are in at least first normal form [Date, 2012]. Therefore, no relation may contain multiple values in a row (and domain) of an attribute of the relation. To accord with the formalism of MEBN, we introduce a new kind of normalization for MEBN-RM, which we call *Entity-Relationship Normalization*.

An MTheory developed from an RDB represents entities. We would like to derive these entities from the RDB. We can do this by defining an entity type in MEBN for each entity relation. This entity type can then be referenced in another relation by using the primary key of the entity relation as a foreign key in a referring relation. For example, we can identify an entity of type *Vehicle* corresponding to the *Vehicle* relation from Fig. 2, and use the primary key ***VehicleID*** to refer to a specific vehicle instance.

In order for this method to produce a clearly defined mapping, we must make sure that all entity types we wish to represent in the MEBN model are represented as entity relations. As an example of a problem that can occur if this practice is not followed, consider an example of a relationship relation that contains a primary key consisting of two attributes that are not foreign keys. For example, we might represent patrol assignments using a *PatrolAssignment* relation having attributes ***PatrolDriver***, ***PatrolNavigator***, PatrolVehicle/Vehicle, and PatrolRegion/Region. The latter two attributes, the vehicle used and the region patrolled, are foreign keys pointing to the *Vehicle* and *Region* relations, respectively. The first two refer to the driver and navigator. These refer to soldiers. If we used this two-attribute primary key to define an entity type, we might erroneously create two different types, when the intention was that both would be filled by an entity of type *Soldier*. To address this issue, we would create a *Soldier* relation with its own original primary key (Definition 2.10), and redefine *PatrolDriver* and *PatrolNavigator* as foreign keys pointing to the *Soldier* relation (i.e., ***PatrolDriver/Soldier***, ***PatrolNavigator/Soldier***, PatrolVehicle/Vehicle, and PatrolRegion/Region).

To formalize this idea, we define *Entity-Relationship Normalization* to ensure that each entity instance is uniquely identified and to clarify which attributes in a relation correspond to entities in MEBN.

**Definition 3.1 (Entity-Relationship Normalization)** A relation is in *Entity-Relationship Normal Form* if either it is an entity relation schema in Definition 2.12 or a relationship relation schema in Definition 2.13.

In the example of Fig. 2, the relation *VehicleLocation* contains a primary key consisting of the attributes *LocatingVehicleID* and *LocatingTimeID*. *LocatingVehicleID* is a foreign key, while *LocatingTimeID* is not. Therefore, in Entity-Relationship normalization, a new relation for *LocatingTimeID* should be added (e.g., the relation *Time*) and the attribute *LocatingTimeID* should be changed to a foreign key pointing to the new relation *Time*. As a result of this transformation, there are three relations (*Vehicle*, *Region*, and *Time*) in which a primary key for each of them consists of a single attribute. These relations are used to identify entities in a MEBN model. Thus, there are three entities; *Vehicle, Region,* and *Time*.

MEBN-RM provides a conversion from a relation schema (RS) in Entity-Relation Normal Form to a *partial MFrag* containing a set of context and resident nodes. Full conversion from a relation instance to a *complete MFrag* (i.e., context nodes, resident nodes, input nodes, a directed acyclic graph, and local distributions) requires augmenting MEBN-RM with either a human modeler or a machine learning algorithm. Hence, in the following sections, MFrag should be taken to mean a partial MFrag.

### *5.1. Entity Mapping*

In MEBN, an entity is a unique kind of thing which exists distinctly and independently, and can be instantiated as an object in the world. For example, from a person entity, various person instances can be defined (e.g., John and Mathew). In RM, an entity relation, a relation containing an original primary key consisting of exactly one attribute, represents a kind of thing that exists uniquely and independently. In MEBN-RM, an entity relation or a non-foreign-key attribute can be mapped to an entity in MEBN as defined by the following.

**Definition 3.2 (ERS to Entity Mapping)** An *ERS to entity mapping* is a mapping in which an entity relation schema, ERS, in an RDBS in Entity-Relationship Normal Form is mapped to an entity, E, denoted by ERS ↦ E.[‡]

For example, the entity relation schema (Definition 2.12) *Vehicle* can be mapped to an entity VEHICLE. In MEBN-RM, entities are written as strings of uppercase letters.

### *5.2. Resident Node Mapping*

In MFrags, a resident node can be described as *Function* or *Predicate* of FOL. MEBN allows the modeler to specify a probability distribution for the truth-value of a predicate or the value of a function. Formulas are not probabilistic and are defined by built-in MFrags [Laskey, 2008]. In this section, we describe the correspondence between functions and predicates in FOL and relations in RM.

[‡] A ↦ B means A is mapped to B.

Table 1. Resident node types on MEBN-RM

| Type | Name | Example |
|---|---|---|
| 1 | Predicate | Follow(*followingvehicleid, leadingvehicleid*) |
| 2 | Function | VehicleClass(*vehicleid*), TerrainType(*regionid*), ContainingRegion(*regionid*), Location(*locatingvehicleid*, *locatingtimeid*) |

Table 1 shows the two types of the resident node with examples from the RDBS *Vehicle Identification*. These are discussed in the next subsection.

***A) Predicate***

In FOL, a predicate represents a true/false statement about entities in the domain. It is expressed by a predicate symbol followed by a list of arguments. For example, Follow(*x*, *y*) is a predicate that expresses whether a following vehicle indicated by the argument *x* is following a leading vehicle indicated by the argument *y*.

In MEBN, this predicate corresponds to a Boolean RV with possible values *true* and *false*. In RM, we can express a predicate as a relation schema in which the attributes are arguments of the predicate, and the rows of the table represent the arguments for which the predicate is true [Date, 2012]. For example, the relation Follow [***FollowingVehicleID***, ***LeadingVehicleID***] can be mapped to a predicate, Follow (*followingvehicleid*, *leadingvehicleid*). The arguments of this predicate are identical to the set of attributes of the relation to which the predicate refers. A predicate from a relation can map to only a true value, because RM doesn't provide a false value for the predicate. For example, suppose that there is a dataset for the relation Follow({{*v1*, *v2*}, {*v2*, *v3*}}). This dataset can be a mapped to a set of propositions of the predicate ({Follow(*v1*, *v2*) = *true*, Follow(*v2*, *v3*) = *true*}).

Table 2 defines the relationship between elements of RM and elements of MEBN for a predicate.

Table 2. Predicate mapping in MEBN-RM

| RM | MEBN |
|---|---|
| Name of relation | Name of Predicate |
| Key | Arguments for Predicate |
| Presence of a tuple | *true* value |
| Absence of a tuple | *false* value |

The name of a relation is used for the name of the predicate corresponding to the relation. The attributes of the relation correspond to the arguments of the predicate in sequence. A given tuple can be either present or absent in the RDB. If the tuple is present, a true value for the corresponding predicate can be assigned in the MEBN representation. If the tuple is absent, a false value for the corresponding predicate can be assigned in the MEBN representation.[§] Now, we introduce a predicate resident node mapping.

[§] This convention is used when adopting the closed world assumption, which asserts that all positive cases of a relation are represented in the database, so that absence of an instance implies the corresponding predicate is false. Dropping the closed world assumption could be handled by adding a *Truth Value* attribute, with values *True* and *False*. With this representation, all cases not appearing in the database would have unknown truth-value.

**Definition 3.3 (Predicate Resident Node Mapping)** A *predicate resident node mapping* is a mapping in which a primary key, ***PK***, of a relationship relation schema, RRS[***PK***], is mapped to a resident node, R, denoted by RRS[***PK***] ↦ R[RRS($A_1$, $A_2$, ..., $A_n$)] = {*true*, *false*}, where ***PK*** = {$K_1$, $K_2$, ..., $K_n$} and $K_i$↦ $A_i$.

For example, the relation schema Follow [***FollowingVehicleID***, ***LeadingVehicleID***] can be mapped to a *predicate resident node* denoted by R[Follow (*followingvehicleid*, *leadingvehicleid*))].

***B) Function***

In FOL, a function is a mapping from domain entities called inputs to a value called the output. For example, the function VehicleClass(*vehicleid*) is a function that maps its argument to *wheeled* if it is a wheeled vehicle and *tracked* if it is a tracked vehicle. In RM, a function is represented by a Non-Foreign-Key Attribute (NF) or Non-Primary Foreign Key (NK) of a relation, because both functionally depend on a Primary Key (PK). Thus, a function of a relation maps to its argument(*s*), the primary key(*s*) for the relation, to the output, which is the value of the domain of the attribute in the relation.

Table 3. Function mapping in MEBN-RM

| NF or NK of RM | Resident Node of MEBN |
|---|---|
| Non-Foreign-Key Attribute/ Non-Primary Foreign Key | Function |
| Primary Key | Arguments of Function |
| Domain of Attribute | Domain of Function |

Table 3 defines the relationship between elements of RM and elements of MEBN for a function. We define a mapping between an element of A or NK of RM, and a function of a resident node of MEBN formally.

**Definition 3.4 (Function Resident Node Mapping)** A *function resident node mapping* is a mapping in which an attribute, A, of a relation schema, RS, and a primary key, **PK**, of the RS is mapped to a resident node, R, of an MFrag, denoted by RS[***PK***, A] ↦ R[A($K_1$, $K_2$, ..., $K_n$)] = Dom(A), where ***PK*** = {$K_1$, $K_2$, ..., $K_n$}.

For example, the argument of the function VehicleClass(*vehicleid*) is the primary key of the relation *Vehicle*, and the output is the value (either *tracked* or *wheeled*) of the attribute *VehicleClass*. In other words, Vehicle[***VehicleID***, *VehicleClass*] ↦ R[VehicleClass(*vehicleid*)] = Dom(*VehicleClass*).

### *5.3. Relation Schema and MFrag Mapping*

In the previous section, we discussed the mapping between the elements of the relation schema and the elements of the MFrag. In this section, we discuss the mapping between a relation schema and a partial MFrag. It is called *RS-MFrag Mapping*. For ERS and RRS, we define the RS-MFrag mapping formally.

**Definition 3.5 (RS-MFrag Mapping)** An *RS-MFrag Mapping* is a mapping in which a relation schema, RS is mapped to a partial MFrag, F, denoted by RS[***PK***, ***O***] ↦ F[***C***, ***R***]. Here, ***C*** denotes a set of context nodes (Definition 2.1), ***R*** denotes a set of resident nodes (Definition 2.1), ***PK*** = {$K_1$, $K_2$, ..., $K_n$} is the primary key, and ***O*** = {$O_1$, $O_2$, ..., $O_m$}, where $O_i$ is the i-th NF or NK attribute. The mapping satisfies the following conditions:

a) If the RS is an ERS and |***O***| > 0, the ***PK*** and ***O*** of the ERS are mapped to the ***C*** and ***R*** of the F, respectively. This is denoted by ERS[***PK***, ***O***] ↦ F[$C_1$[IsA($K_1$, E($K_1$))], $R_1$[$O_1$($K_1$)], …, $R_m$[$O_m$($K_1$)]].

b) If the RS is an RRS and |***O***| > 0, the ***PK*** and ***O*** of the RRS are mapped to the ***C*** and ***R*** of the F, respectively. This is denoted by RRS[***PK***, ***O***] ↦ F[$C_1$[IsA($K_1$, E($K_1$))], …, $C_n$[IsA($K_n$, E($K_n$))], $R_1$[$O_1$(***PK***)],…, $R_m$[$O_m$(***PK***)]]**.
c) If the RS is an RRS and |***O***| = 0, the ***PK*** and RRS are mapped to the ***C*** and ***R*** of the F, respectively. This is denoted by RRS[***PK***] ↦ F[$C_1$[IsA($K_1$, E($K_1$))], …, $C_n$[IsA($K_n$, E($K_n$))],$R_1$[RRS(***PK***)]].

Case (a) is that the RS is an ERS and it has at least one attribute which is not used for the primary key. In this case, the single attribute $K_1$ in ***PK*** is used to create the IsA context node, and each attribute in ***O*** is mapped to each resident node respectively using the function resident node mapping. For example, the relation Vehicle[***VehicleID/Vehicle***, VehicleClass] becomes a partial MFrag denoted by F[C[IsA(*vehicleid*, VEHICLE)], R[VehicleClass(*vehicleid*)]].

Case (b) is that the RS is an RRS and it has an attribute which is not used for the primary key. Each attribute $K_i$ in ***PK*** is used to create its respective IsA context node and each attribute in ***O*** is mapped to a resident node using the function resident node mapping. For example, the relation VehicleLocation[***LocatingVehicleID/Vehicle***, ***LocatingTimeID/Time***, Location/Region] becomes a partial MFrag denoted by F[C[IsA(*locatingvehicleid*, VEHICLE), IsA(*locatingtimeid*, TIME), R[Location(*locatingvehicleid*, *locatingtimeid*)]].

Case (c) is that the RS is the RRS and has no attributes other than the primary key. In this case, each attribute $K_i$ in ***PK*** is used to create the IsA context node and the relation is mapped to a predicate resident node, RRS($K_1$, $K_2$, …, $K_n$), using the *Predicate resident node Mapping*. For example, the relation Follow [***FollowingVehicleID/Vehicle***, ***LeadingVehicleID/Vehicle***] becomes a partial MFrag denoted by F[C[IsA(*followingvehicleid*, VEHICLE), IsA(*leadingvehicleid*, VEHICLE), R[Follow(*followingvehicleid*, *leadingvehicleid*)]].

### *5.4. Relational Database Schema and MTheory Mapping*

In the previous section, we discussed the mapping between a relation schema and partial MFrag. In this section, we discuss the mapping between a relational database schema (RDBS) and MTheory (M). It is called *RDBS-MTheory Mapping*. Basically, the mapping produces one MTheory from one relational database schema, denoted by the following.

**Definition 3.6 (RDBS-MTheory Mapping)** An*RDBS-MTheory Mapping* is a mapping in which a relational database schema RDBS is mapped to an MTheory M (Definition 2.3), denoted by RDBS[$RS_1$, $RS_2$, …, $RS_n$] ↦ M[$F_1, F_2, \ldots, F_n$], where $RS_i$ is a relation schema in the RDBS, $F_i$ is a partial MFrag in the M, and n is the number of the relation schemas in the RDBS and the number of the partial MFrags in the M, if the *RS-MFrag Mapping* between $RS_i$ and $F_i$ is able to be used.

For example, the *Vehicle Identification* RDBS can be directly an MTheory using the RDBS-MTheory mapping. The relations *Vehicle*, *Region*, and *VehicleLocation* in the RDBS are converted to partial MFrags *Vehicle*, *Region*, and *VehicleLocation*. The following section presents a mapping algorithm using MEBN-RM, which is a process to develop an MTheory from data in RM and contains specific sub-steps.

### *5.5. MEBN-RM Mapping Algorithm*

In the previous sections, we discussed the mapping definitions for entities, resident nodes, MFrags, and MTheories. This section presents a *MEBN-RM mapping algorithm* (Algorithm 1) which performs the *RDBS-MTheory Mapping* in Definition 3.6 and specifies how to convert an MTheory from a relational database

**E(X) is the entity type which the attribute X points to.

schema using the MEBN-RM definitions.

For the *MEBN-RM mapping algorithm*, we assume that (1) the relational database schema are normalized by *Entity-Relationship Normalization* (Definition 3.1), and (2) the list of relation schemas in the relational database schema are sorted by the entity relation schemas (ERS) first and the relationship relation schemas (RRS) second. For the algorithm, let *M* be an MTheory, *M*.***E*** be a set of entity types of *M*, *M*.***F*** be a set of MFrags, *F*.***C*** be a set of context nodes in an MFrag *F*, *F*.***R*** be a set of resident nodes of *F*, *RDBS* be a relational database schema, *rs* be a relation schema in *RDBS*, and *rs*.***O*** be a set of attributes for NF and/or NK of *rs*. The algorithm takes the relational database schema *RDBS* as an input and produces the MTheory *M* as an output.

**Algorithm 1**: MEBN-RM Mapping

```
Input: RDBS a relational database schema
Output: M a partial mapped MTheory
Procedure MEBN-RM_Mapping ( RDBS )
1   M ← create a default MTheory
2   M.name ← get a schema name using RDBS
3    for i = 1, … until n
4    rs_i ← get i-threlation schema from RDBS
5    RM-MFrag_Mapping(rs_i, M)
6    return M

Procedure RM-MFrag_Mapping ( rs,   // relation schema
                             M     // default MTheory
          )
7    if rs = ERS then
8      M.E ← create an entity type from rs using ERS to Entity Mapping
9    if rs = ERS & |rs.O| > 0 then
10     M.F ← F ← create an MFrag for rs
11     F.C ← create IsA nodes from the entity types M.E associated with M.F
12     F.R ← create resident nodes from rs using Function resident node Mapping
13   else if rs = RRS then
14     M.F ← F ← create an MFrag for rs
15     F.C ← create IsA nodes from the entity types M.E associated with M.F
16     if |rs.O| = 0 then
17        F.R ← create resident nodes from rs using Predicate resident node Mapping
18     else if |rs.O| > 0 then
19        F.R ← create resident nodes from rs using Function resident node Mapping
20   end
```

Inputs of this algorithm are a relational database schema *RDBS*. (1) The algorithm starts with creating a default MTheory *M*. (2) The name of *RDBS* is used to create the name of *M*. (3)(4) All relation schema are investigated from a first relation schema $rs_1$ to a last relation schema $rs_n$, where *n* denotes the number of the relation schemas in *RDBS*. (5) For an *i*-th relation schema, the algorithm performs the procedure *RM-MFrag Mapping* defined in Definition 3.5. (7) If the *i*-th relation schema is ERS, (8) the *ERS to Entity Mapping* (Definition 3.2) is performed. (9) If the *i*-th relation schema *rs* is ERS and there is an attribute O for NF or NK, then (10) an MFrag

*F* for the *rs* is created and added to the set of MFrags of *M*, (11) IsA context nodes are created from the entity types ***M.E*** associated with *F* and added to the set of context nodes of *F*, and (12) resident nodes are created from *rs* using the *Function resident node Mapping* (Definition 3.4) and added into the set of resident nodes of *F*. (13) If the *i*-th relation schema *rs* the is RRS, then performs (14) to (19). (14) An MFrag *F* for the *rs* is created and added to the set of MFrags of *M*. (15) IsA context nodes are created from the entity types ***M.E*** associated with *F* and added to the set of context nodes of *F*. (16) If there is no attribute O for NF or NK, (17) the *Predicate resident node Mapping* (Definition 3.3) for *rs* is performed.(18) If there is an attribute O for NF or NK, (19) the *Function resident node Mapping* (Definition 3.4) for *rs* is performed. (6) The algorithm results in the MTheory *M*. We consider the complexity of this algorithm in terms of the Big O. The for-loop in Line 3 is the most computationally intensive operation of this algorithm and it is influenced by the number of the relations *n*. The complexity of the procedure *RM-MFrag_Mapping* is O(*k*), where *k* is the number of primary keys in a relation. Therefore, this algorithm has complexity O(*nk*). However, in the most cases, *k* is not a big number (i.e., $k < n$), so we can consider this algorithm as O(*n*). We will see experiment results for this in Section 4.2.

MTheory 2 shows a result for the RDBS 1 using the MEBN-RM Mapping algorithm.

**MTheory 2**: Vehicle Identification

1 [F: Vehicle
2 [C: IsA(*vehicleid*, VEHICLE)]
3 [R: VehicleClass(*vehicleid*)]
4 ]
5 [F: Region
6 [C: IsA(*regionid*, REGION)]
7 [R: TerrainType(*regionid*)]
8 [R: ContainingRegion(*regionid*)]
9 ]
10 [F: VehicleLocation
11 [C: IsA(*locatingvehicleid*, VEHICLE)]
12 [C: IsA(*locatingtimeid*, TIME)]
13 [R: Location(*locatingvehicleid*, *locatingtimeid*)]
14 ]
15 [F: Follow
16 [C: IsA(*followingvehicleid*, VEHICLE)]
17 [C: IsA(*leadingvehicleid*, VEHICLE)]
18 [R: Follow(*followingvehicleid*, *leadingvehicleid*)]
19 ]

For example, the relation *Region* in RDBS 1 is an ERS, which is mapped to the entity REGION and used to create the IsA context node in Line 6. The relation *Region* contains an attribute *TerrainType* which is converted to a resident node TerrainType as a Function in Line 7. Also, the relation *Region* contains a Non-Primary Foreign Key *ContainingRegion* which is mapped to a resident node ContainingRegion as a Function in Line 8.

## 6. Experiment for MEBN-RM

In this section, we present an experiment to evaluate the performance of the MEBN-RM algorithm in terms of mapping speed and quality. The MEBN-RM algorithm was implemented on an open-source program. First this program is introduced and then the experiment is presented.

### *6.1. MEBN-RM Tool*

MEBN-RM Tool is a JAVA based open-source program[††] that can be used to create an MTheory script from a relational schema. MEBN-RM Tool can be commonly used in MEBN learning or MEBN modelling. MEBN-RM Tool is implemented in the MEBN-RM mapping algorithm in Section 3.5. This enables rapid development of an MTheory script by just clicking a button in the tool. The current version of MEBN-RM Tool uses MySQL, an open-source relational database management system, to take the relational schema. The most recent version and source code of MEBN-RM Tool are available online at the GMU_HMLP Github repository (https://github.com/pcyoung75/GMU_HMLP.git).[‡‡]

Once we obtain MEBN-RM Tool we are ready to select a relational database and convert it to an MTheory script. MEBN-RM Tool contains two panels: (1) a left tree panel shows a list of relational databases and (2) a right panel shows a result MTheory script. By selecting a database and clicking the select button in the tool, the MEBN-RM mapping performs and produces a result MTheory script.

### *6.2. Experiment*

We conducted the experiment to evaluate the performance of the MEBN-RM algorithm in terms of the mapping time and accuracy. The mapping time is the time it takes to map from a relational database to an MTheory script. The mapping accuracy means how correctly the MTheory script was mapped from the relational database. For this, we compared both elements from the MTheory script and the relational database.

For the test relational databases, Relational Learning Repository [Motl & Schulte, 2015], which contains more than 70 relational databases from the real world or the simulation, was used. For the experiment, 10 real world relational databases (see Table 4) from 8 domains (Education, Financial, Entertainment, Government, Industry, Kinship, Medicine, and Social) were chosen. These relational databases satisfied *Entity-Relationship Normalization* (Definition 3.1), so the normalization step was not required. The experiment was run on a 3.40GHz Intel Core i7-3770 processor.

#### *A) Mapping Time*

Table 4 shows 10 relational databases with the name, domain, and number of attributes/relations. Each relational database had different attribute and relation features, so the following factors were used to investigate the mapping time: (1) the number of relations and (2) the number of attributes.

Table 4 also shows the experiment results for the mapping time over the different number of attributes and relations in each of the 10 real world relational databases. The correlation coefficient for the mapping time over the number of attributes was -0.033, while the correlation coefficient for the mapping time over the number of relations was 0.97 (see Figure 3).

[††] Researchers around the world can debug and extend the MEBN-RM Tool.
[‡‡] Github is a distributed version control system (https://github.com).

Table 4. 10 real world relational databases with experiment results for the mapping time

| # | Name | Domain | # of RS (Def. 2.12) | # of ERS (Def. 2.13) | # of RRS (Def. 2.14) | # of Attributes | # of Primary Keys (Def. 2.8) | Mapping Time (Sec.) |
|---|---|---|---|---|---|---|---|---|
| 1 | Stats | Education | 8 | 8 | 0 | 71 | 8 | 0.0597 |
| 2 | Financial | Financial | 8 | 8 | 0 | 55 | 8 | 0.0498 |
| 3 | MovieLens | Entertainment | 7 | 4 | 3 | 24 | 10 | 0.0445 |
| 4 | LegalActs | Government | 5 | 2 | 3 | 33 | 7 | 0.0334 |
| 5 | SAT | Industry | 36 | 3 | 33 | 69 | 37 | 0.1656 |
| 6 | Dunur | Kinship | 17 | 1 | 16 | 34 | 33 | 0.0726 |
| 7 | Elti | Kinship | 11 | 1 | 10 | 22 | 21 | 0.0503 |
| 8 | Bupa | Medicine | 9 | 2 | 7 | 16 | 9 | 0.0383 |
| 9 | Pima | Medicine | 9 | 1 | 8 | 18 | 9 | 0.0417 |
| 10 | Facebook | Social | 2 | 1 | 1 | 265 | 3 | 0.0359 |

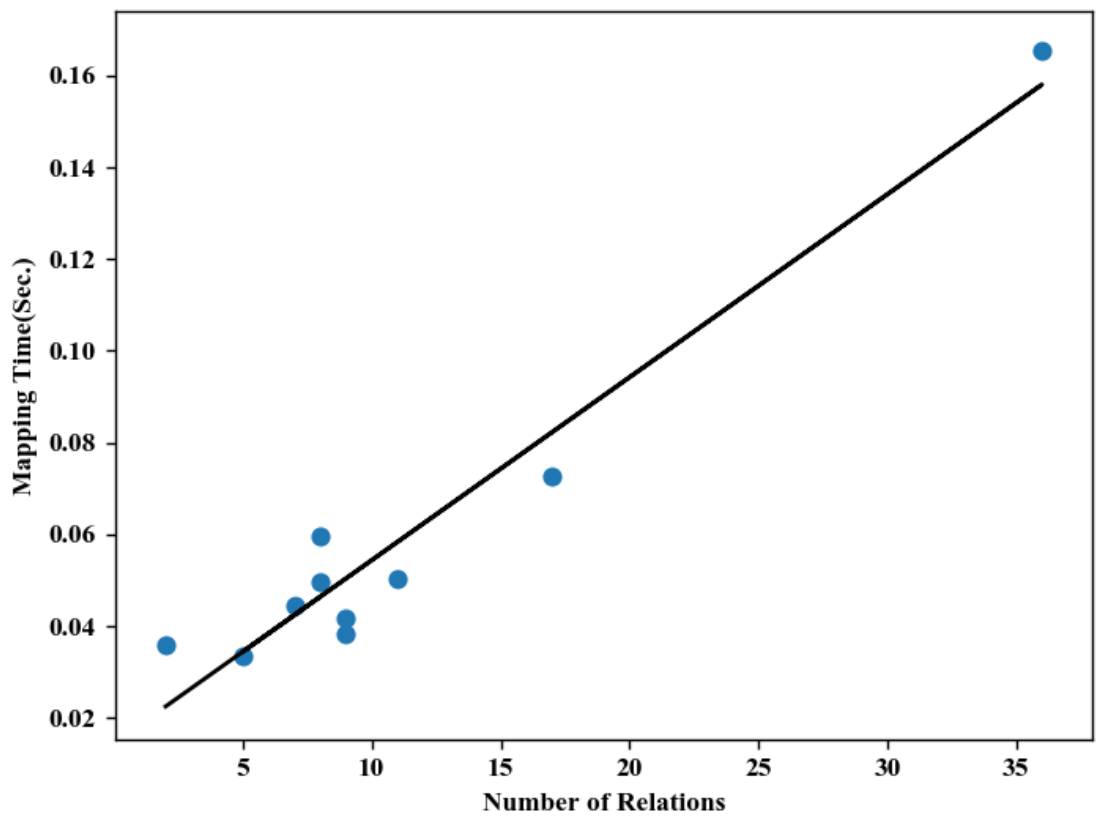

Figure 3. Mapping time over the number of relations

The increase in mapping time was linear in the number of relations. However, the acceptable speed of the algorithm alone is not sufficient. The accuracy for the mapped results is also essential.

### *B) Mapping Accuracy*

The mapping accuracy for the MEBN-RM algorithm can be evaluated by comparing between the numbers of elements in a source RDB and the numbers of elements in a target MTheory. For example, an ERS (Definition 2.12) in the RDB and an entity in the MTheory are mapped to one-to-one. An ERS containing NK (Definition 2.8) or A (Definition 2.9) is mapped to an MFrag. Table 5 shows the numbers of elements in the target MTheories. These elements were generated as we expected.

Table 5. 10 real world relational databases with elements of its mapped MTheory

| # | Name | # of Entity | # of MFrag (Def. 2.1) | # of Resident Node (Def. 2.2) | # of IsA Nodes (Def. 2.5) |
|---|---|---|---|---|---|
| 1 | Stats | 8 | 8 | 63 | 8 |
| 2 | Financial | 8 | 8 | 47 | 8 |
| 3 | MovieLens | 4 | 7 | 14 | 10 |
| 4 | LegalActs | 2 | 5 | 28 | 7 |
| 5 | SAT | 3 | 33 | 33 | 34 |
| 6 | Dunur | 1 | 16 | 16 | 32 |
| 7 | Elti | 1 | 10 | 10 | 20 |
| 8 | Bupa | 2 | 7 | 7 | 7 |
| 9 | Pima | 1 | 9 | 9 | 9 |
| 10 | Facebook | 1 | 2 | 263 | 3 |

We also applied MEBN-RM to two projects. Section 5 introduces the projects with specific relational databases and mapping outcomes.

## 7. Use Cases

In this section, we introduce two example use cases using MEBN-RM: a Critical Infrastructure Defense System and a Smart Manufacturing System.

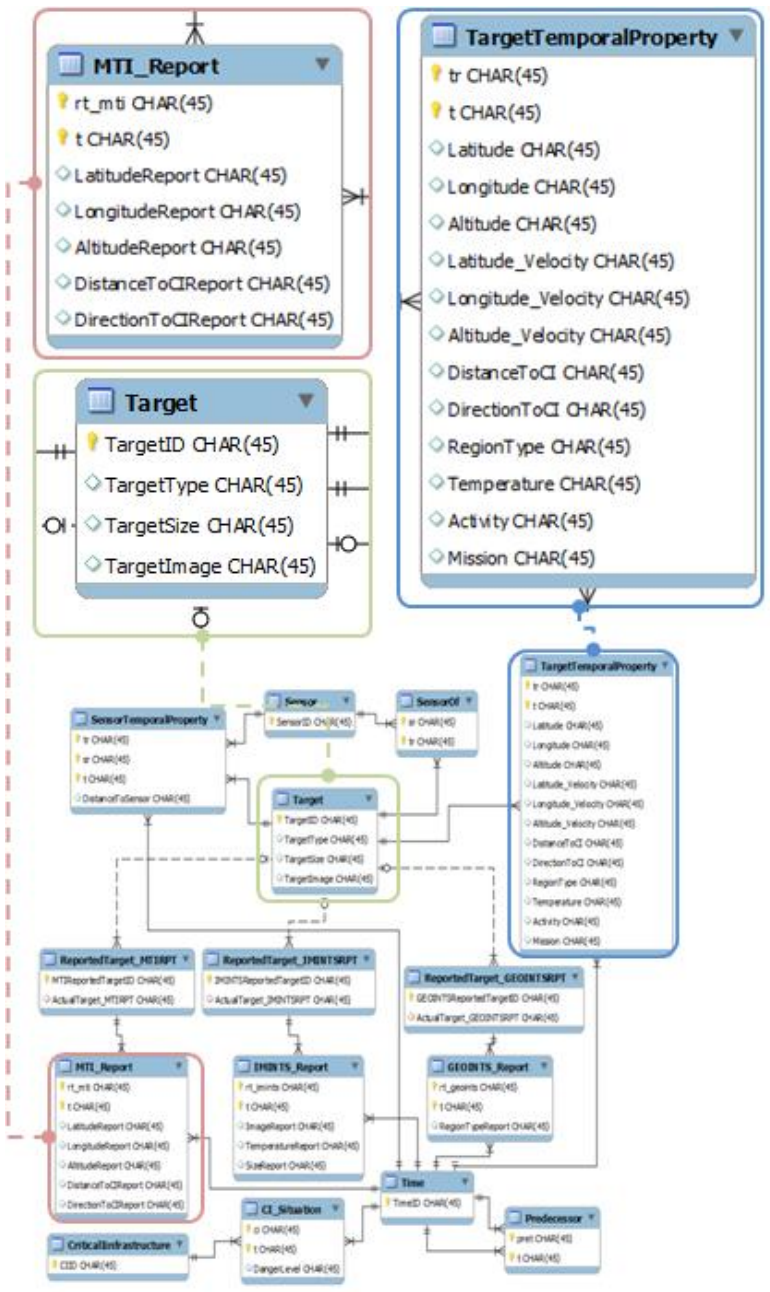

Figure 4. Partial HERALD Relational Database Schema

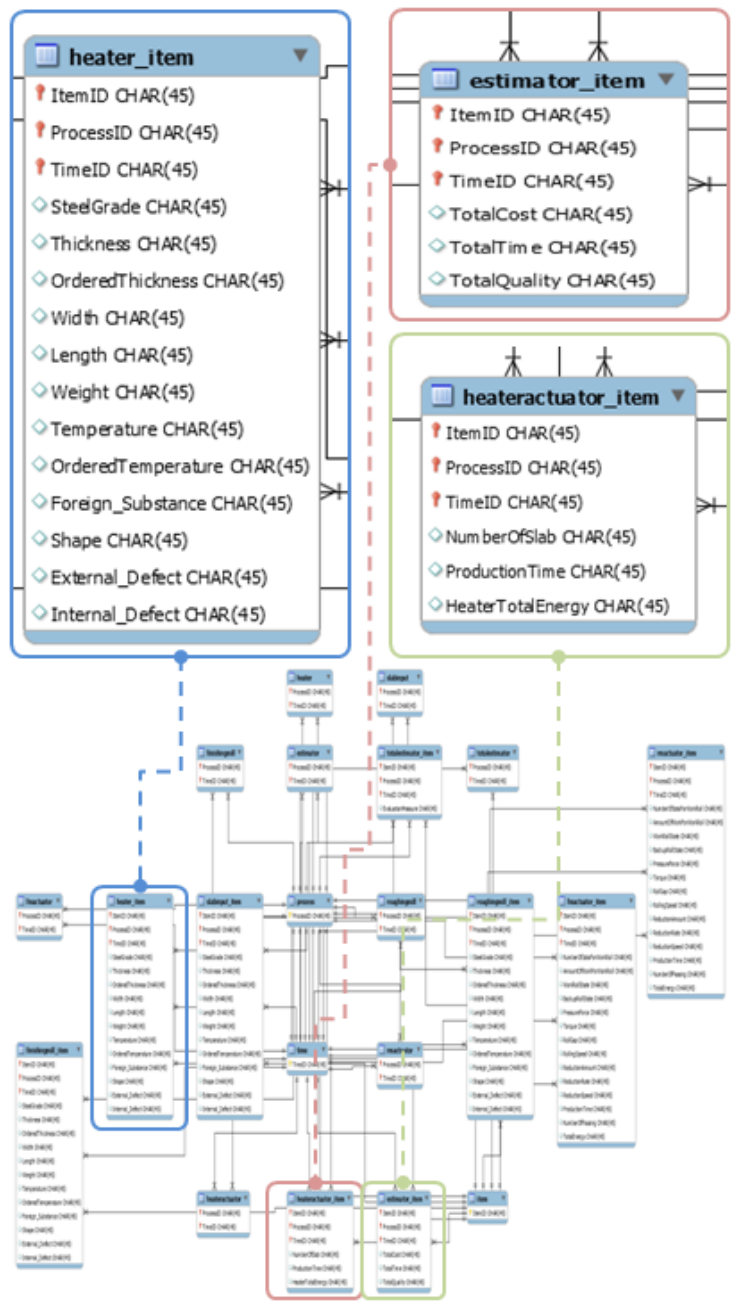

Figure 5. Partial MSAW Relational Database Schema

### *7.1. Critical Infrastructure Defense System*

HERALD is a proof-of-concept system designed to prevent attacks to critical infrastructures (CI) through early detection/identification of threatening targets and short-term prediction of the target's activities and threat level for the region where the critical infrastructures are located [Park & Laskey, 2016]. In 2014, Samsung Thales coined the design of a critical infrastructure defense system which is the predecessor of the HERALD system. Requirements from Samsung Thales were to design and develop a next-generation system for critical infrastructure defense by means of integration for previous knowledge with evidence from multiple sensors (e.g., MTI (Moving Target Indicator) system, IMINT (Imagery Intelligence) sensor system, and GEOINT (Geospatial Intelligence) system).

HERALD consisted of an inference module, a control module and a scenario simulator. The inference module used a HERALD MTheory to infer current and future situations. The HERALD MTheory was designed by domain experts and learned using a MEBN learning process (HMLP) [Park et al., 2016]. Also, HERALD contained a relational database that was developed by the domain experts and used for a simulation. The HERALD simulator simulated ground truth information of a situation in which our forces and enemies operated against each other. To develop the HERALD MTheory, the relational database was used to develop a partial HERALD MTheory. Fig. 4 shows the schema of the relational database.

The relational database schema contained 18 relations (e.g., MTI_Report, Target, and TargetTemporalProperty). The relation MTI_Report represented knowledge from MTI about reported location (e.g., LatitudeReport) of and reported moving aspects (e.g., DistanceToCIReport) of a target. The relation Target represented knowledge about a target's information that was not time-varying. The relation TargetTemporalProperty represented knowledge about a target's information that was time-varying.

MEBN-RM was used to convert the relational database schema to the partial HERALD MTheory. MTheory 3 shows some MFrags of the partial HERALD MTheory. These three MFrags correspond to the three relations in the relational database schema. Note that each name of the resident nodes is changed to a form that includes an abbreviation derived from the name of the MFrag (e.g., the prefix MR for resident nodes in the MFrag MTI_Reprt) to prevent construction for resident nodes with the same name. The domain experts and the MEBN learning process, then, used the partial HERALD MTheory to construct a complete HERALD MTheory by adding class local distributions for resident nodes and conditional dependence relationships between the resident nodes.

**MTheory 3**: Part of Script MTheory for HERALD

```
[F: MTI_Report
    [C: IsA(rt_mti, REPORTEDTARGET_MTIRPT)]
    [C: IsA(t, TIME)]
    [R: MR_LatitudeReport(rt_mti, t)]
    [R: MR_LogitudeReport(rt_mti, t)]
    [R: MR_AltitudeReport(rt_mti, t)]
    [R: MR_DistanceToCIReport (rt_mti, t)]
    [R: MR_DirectionToCIReport(rt_mti, t)]
]
[F: target
    [C: IsA(targetid, TARGET)]
    [R: T_TargetType(targetid)]
    [R: T_TargetSize(targetid)]
    [R: T_TargetImage(targetid)]
]
…
[F: TargetTemporalProperty
```

```
    [C: IsA(tr, TARGET)]
    [C: IsA(t, TIME)]
    [R: TTP_Latitude(tr,t)]
    [R: TTP_Longitude(tr,t)]
    [R: TTP_Altitude(tr,t)]
    [R: TTP_Latitude_Velocity(tr,t)]
    [R: TTP_Longitude_Velocity (tr,t)]
    [R: TTP_Altitude_Velocity (tr,t)]
    [R: TTP_DistanceToCI(tr,t)]
    [R: TTP_DirectionToCI(tr,t)]
    [R: TTP_RegionType(tr,t)]
    [R: TTP_Temperature(tr,t)]
    [R: TTP_Activity(tr,t)]
    [R: TTP_Mission(tr,t)]
]
```

### *7.2. Smart Manufacturing System*

An MSAW (Predictive Manufacturing Situation Awareness) system as a prototype system was designed and developed to estimate current situations as well as predict future situations for a steel plate manufacturing [Park et al., 2017].The MSAW system was associated with various equipments for steel plate manufacturing (e.g., a reheating furnace, a roughing mill, and a finishing mill) to produce steel plates of good quality (e.g., few defects and required flatness) by taking steel slabs. The goal of the MSAW system was to support four smart functions: Function 1 (Control value reasoning given outputs), Function 2 (Optimal control value finding to maximize/minimize objective values (e.g., outputs)), Function 3 (Prediction for future manufacturing situations), and Function 4 (Sensitivity analysis to find defective factors for faulty outputs). The MSAW system supporting the steel plate manufacturing contained an MSAW MTheory which was used to perform the four functions.

The MSAW MTheory was designed by domain experts and learned using data from a simulator. The simulator was designed by domain experts and contained a relational database as shown Fig. 5. The simulator simulated ground truth information for the reheating furnace, the roughing mill, and the finishing mill. The relational database in the simulator was used to develop a partial MSAW MTheory. Fig. 5 shows the schema of the relational database.

The relational database schema contained 21 relations (e.g., heater_item, estimator_item, and heateractuator_item). The relation heater_item represented properties of a slab item heated by the reheating furnace. The relation contained several attributes for the slab item (e.g., attributes SteelGrade, Thickness, and Temperature). The relation estimator_item represented knowledge about an overall situation for manufacturing in terms of total manufacturing cost, total manufacturing time, and total quality rate for products. The relation heateractuator_item represented knowledge about control factors and properties for the reheating furnace (e.g., attributes NumberOfSlab and ProductionTime).

MEBN-RM was used to convert the relational database schema to the partial MSAW MTheory. MTheory 4 shows some MFrags of the partial MSAW MTheory. These three MFrags correspond to the three relations (i.e., heater_item, estimator_item, and heateractuator_item) in the relational database schema. The domain experts and the MEBN learning process, then, used the partial MSAW MTheory to construct a complete MSAW MTheory by adding local probability distributions for resident nodes and conditional dependence relationships between the resident nodes.

**MTheory 4**: Part of Script MTheory for MSAW

```
[F: heater_item
      [C: IsA(itemid, ITEM)]
      [C: IsA(processid, PROCESS)]
      [C: IsA(timeid, TIME)]
      [R: HI_SteelGrade(itemid, processid, timeid)]
      [R: HI_Thickness(itemid, processid, timeid)]
      [R: HI_OrderedThickness(itemid, processid, timeid)]
      [R: HI_Width(itemid, processid, timeid)]
      [R: HI_Length(itemid, processid, timeid)]
      [R: HI_Weight(itemid, processid, timeid)]
      [R: HI_Temperature(itemid, processid, timeid)]
      [R: HI_OrderedTemperature(itemid, processid, timeid)]
      [R: HI_Foreign_Substance(itemid, processid, timeid)]
      [R: HI_Shape(itemid, processid, timeid)]
      [R: HI_External_Defect(itemid, processid, timeid)]
      [R: HI_Internal_Defect(itemid, processid, timeid)]
]
[F: estimator_item
      [C: IsA(itemid, ITEM)]
      [C: IsA(processid, PROCESS)]
      [C: IsA(timeid, TIME)]
      [R: ETMOI_TotalCost(itemid, processid, timeid)]
      [R: ETMOI_TotalTime(itemid, processid, timeid)]
      [R: ETMOI_TotalQuality (itemid, processid, timeid)]
]
…
[F: heateractuator_item
      [C: IsA(itemid, ITEM)]
      [C: IsA(processid, PROCESS)]
      [C: IsA(timeid, TIME)]
      [R: HAI_NumberOfSlab(itemid, processid, timeid)]
      [R: HAI_ProductionTime(itemid, processid, timeid)]
      [R: HAI_HeaterTotalEnergy(itemid, processid, timeid)]
]
```

## 8. Conclusion

In this research, we presented MEBN-RM formalizing conversion from a relational database schema in RM to a partial MTheory in MEBN syntactically. To do this, MEBN-RM contained the four levels of the mappings between elements of the relational database schema and MTheory. Table 6 summarizes the mappings which this research presents.

Table 6. Mapping types on MEBN-RM

| RM | Mapping Types | MEBN |
|---|---|---|
| ERS | Definition 3.2 ERS to Entity Mapping | Entity |
| RRS | Definition 3.3 Predicate resident node Mapping | Predicate resident node |
| Non-foreign-key attribute, non-primary foreign key | Definition 3.4 Function resident node Mapping | Function resident node |
| RS | Definition 3.5 RS-MFrag Mapping | MFrag |
| RDBS | Definition 3.6 RDBS-MTheory Mapping | MTheory |

MEBN-RM is a foundation of designing a MEBN model from a relational database, so, using MEBN-RM, the modeler (Human or Machine) can design the MEBN model seamlessly. The idea behind MEBN-RM may be used to develop other mapping models for different types of database (e.g., ontology, graph, and event database) as an example mapping model. Recently non-relational databases, called *NoSQL*, are receiving increasing attention. In the era of Big Data, we may need a scalable and flexible database to manage the many and varied types of data. In this research, we only focused on the relational model as a source data model to develop an MTheory. Future work will consider extensions to NoSQL data and other types of data.

**Acknowledgements**

The research was partially supported by the Office of Naval Research (ONR), under Contract#: N00173-09-C-4008. We appreciate Dr. Paulo Costa and Mr. Shou Matsumoto for their helpful comments on this research. We also sincerely thank Dr. Shin at Samsung Thales for advice on the HERALD MTheory and simulation.